\documentclass[sigconf]{acmart}
\usepackage{algorithm}
\usepackage{algorithmic}
\usepackage{multirow}
\usepackage{subfigure}

\AtBeginDocument{%
  }

\copyrightyear{2023} 
\acmYear{2023} 
\setcopyright{acmlicensed}\acmConference[CIKM '23]{Proceedings of the 32nd ACM International Conference on Information and Knowledge Management}{October 21--25, 2023}{Birmingham, United Kingdom}
\acmBooktitle{Proceedings of the 32nd ACM International Conference on Information and Knowledge Management (CIKM '23), October 21--25, 2023, Birmingham, United Kingdom}
\acmPrice{15.00}
\acmDOI{10.1145/3583780.3615195}
\acmISBN{979-8-4007-0124-5/23/10}

\begin{document}

\title{Neighborhood Homophily-based Graph Convolutional Network}


% author 1
\author{Shengbo Gong}
\orcid{0000-0001-7482-6524}
\affiliation{%
  \institution{Institute of Cyberspace Security}
  \institution{Zhejiang University of Technology}
  \city{Hangzhou}
  \country{China}
}
\affiliation{%
  \institution{Binjiang Cyberspace Security Institute of ZJUT}
  \city{Hangzhou}
  \country{China}
}
\email{jshmhsb@gmail.com}

% author 2
\author{Jiajun Zhou}
\authornote{Jiajun Zhou is the corresponding author.}
\orcid{0000-0002-5062-4183}
\affiliation{%
  \institution{Institute of Cyberspace Security}
  \institution{Zhejiang University of Technology}
  \city{Hangzhou}
  \country{China}
}
\affiliation{%
  \institution{Binjiang Cyberspace Security Institute of ZJUT}
  \city{Hangzhou}
  \country{China}
}
\email{jjzhou@zjut.edu.cn}

% author 3
\author{Chenxuan Xie}
\orcid{0000-0002-4474-615X}
\affiliation{%
  \institution{Institute of Cyberspace Security}
  \institution{Zhejiang University of Technology}
  \city{Hangzhou}
  \country{China}
}
\affiliation{%
  \institution{Binjiang Cyberspace Security Institute of ZJUT}
  \city{Hangzhou}
  \country{China}
}
\email{hello.crabboss@gmail.com}

% author 4
\author{Qi Xuan}
\orcid{0000-0002-1042-470X}
\affiliation{%
  \institution{Institute of Cyberspace Security}
  \institution{Zhejiang University of Technology}
  \city{Hangzhou}
  \country{China}
}
\affiliation{%
  \institution{Binjiang Cyberspace Security Institute of ZJUT}
  \city{Hangzhou}
  \country{China}
}
\email{xuanqi@zjut.edu.cn}

\renewcommand{\shortauthors}{Shenbo Gong, Jiajun Zhou, Chenxuan Xie, and Qi Xuan}

\begin{abstract}
Graph neural networks (GNNs) have been proved powerful in graph-oriented tasks. However, many real-world graphs are heterophilous, challenging the homophily assumption of classical GNNs. To solve the universality problem, many studies deepen networks or concatenate intermediate representations, which does not inherently change neighbor aggregation and introduces noise. Recent studies propose new metrics to characterize the homophily, but rarely consider the correlation of the proposed metrics and models. In this paper, we first design a new metric, Neighborhood Homophily (\textit{NH}), to measure the label complexity or purity in node neighborhoods. Furthermore, we incorporate the metric into the classical graph convolutional network (GCN) architecture and propose \textbf{N}eighborhood \textbf{H}omophily-based \textbf{G}raph \textbf{C}onvolutional \textbf{N}etwork (\textbf{NHGCN}). 
In this framework, neighbors are grouped by estimated \textit{NH} values and aggregated from different channels, and the resulting node predictions are then used in turn to estimate and update \textit{NH} values.
The two processes of metric estimation and model inference are alternately optimized to achieve better node classification. 
NHGCN achieves top overall performance on both homophilous and heterophilous benchmarks, with an improvement of up to 7.4\% compared to the current SOTA methods. 
\end{abstract}

  



\maketitle

\section{Introduction}
Graph-structured data can effectively model real-world interactive systems and has been widely adopted as a modeling tool in fields such as finance and bioinformatics~\cite{fraudauc,drug}, consequently speeding up the rapid development of graph neural networks (GNNs) for graph-oriented tasks.
Classical GNNs~\cite{kipf2016semi,velivckovic2017gat,hamilton2017inductive} have been proved powerful in analyzing graph datasets that conform to the homophily assumption, where nodes tend to be connected to nodes with the same label. However, many real-world graphs contain heterophily or low homophily, e.g., normal users are usually associated with fraudulent accounts in financial fraud networks.
Such scenarios do not follow the homophily assumption of classical GNNs, making them exposed to too much irrelevant information during message propagation and aggregation, leading to poor performance.

Such cases can be summarized as the universality problem~\cite{chien2020adaptive}, calling for a generic solution applicable to both homophilous and heterophilous graphs, i.e., universality necessitates that models function independently of homophily or heterophily assumptions.

\noindent\textbf{Related Work.}\:
To go beyond the homophily, some spectral methods espouse the intuition that non-local information might be helpful~\cite{liu2021non}. Therefore, they resort to deeper GNNs, since stacking layers is equivalent to larger receptive field sizes, making nodes access to non-local information~\cite{chien2020adaptive}.
But the ensuing problems such as over-smoothing~\cite{li2018deeper,oono2019graph} and model degradation~\cite{zhang2022degradation} need to be alleviated by initial connection, skip connection, and linearization. Representative methods are GCNII~\cite{chen2020gcnii}, GPRGNN~\cite{chien2020adaptive}, BernNet~\cite{he2021bernnet}.
Maintaining local assumptions,
FAGCN~\cite{bo2021fagcn} and ACM~\cite{luan2022acm} design local filters to better capture local information.
\textit{However, limited by the globality of spectral theory, they are inherently unable to group information from different orders and thus introduce noise}~\cite{wang2022hog}.

Spatial methods do not rely on spectral graph theory and thus allow fine-grained adaptations of message propagation rules.
Geom-GCN~\cite{pei2019geom} and WRGAT~\cite{suresh2021wrgat} rewire the graphs to aggregate latent space neighbors or structurally similar neighbors. CPGNN~\cite{zhu2021cpgnn} and HOG~\cite{wang2022hog} reuse the predicted labels to calibrate the message propagation.
\textit{These methods require computing aggregating weights or new topologies and are thus time-consuming.}

Most related studies are GBK\cite{du2022gbk}, GGCN~\cite{yan2022two} and ACM~\cite{luan2022acm}, since they also adopt a three-channel architecture in respective models. 
% GBK groups neighbors by whether they are of the same class as the target nodes and aggregate them adaptively with three channels. But the gating mechanism they used significantly slows down their model. GGCN rewires the graphs by relative degree and then groups neighbors based on the sign of feature similarity. 
However, the first two are not directly from the definition of homophily, though they prove that their associations with homophily on an ad hoc assumption. The difference between ours and ACM is the selection of neighbors rather than low or high frequencies.

\noindent\textbf{Motivation and Contributions.}\:
Most of existing methods ignore the fact that homophilous nodes behave differently with heterophilous nodes. But it can be commonly observed in real-world graphs, e.g., researchers in interdisciplinary studies favor citing literature from multiple fields, while others confine their references within a single discipline~\cite{bornmann2008citation}. 
In such scenarios, mixing messages from ``outward'' and ``inward'' neighbors in the same channel can be considered an unwise practice. Furthermore, the classical metric falls short in common cases~\cite{ma2021necessity}, and new homophily metrics are not well designed for guiding the model.
Therefore, we propose a novel spatial domain approach --- \textbf{N}eighborhood \textbf{H}omophily-based \textbf{G}raph \textbf{C}onvolutional \textbf{N}etwork (\textbf{NHGCN}), in which neighbors are grouped by estimated \textit{NH}, and nodes receive messages from low- and high-\textit{NH} neighbors through different channels. 
\textbf{Our contributions are:} 1) proposing a new metric which can alleviate the dilemma of classical metric and guide the model from end-to-end; 2) devising a simple but effective model with a backbone of GCN; 3) our model achieves SOTA on 7 out of 10 popular datasets containing different level on homophily.
% Fig.~\ref{fig:main} outlines the overall framework of our method.

%%%%%%%%%%%%%%%%%%%%%%%%%%%%%%%%%%%%%%%%%%%%%%%%%%%%%%%%%%%%%%%%%%%%%%%%%%%% section 2 %%%%%%%%%%%%%%%%%%%%%%%%%%%%%%%%%%%%%%%%%%%%%
\section{Preliminaries}
\noindent\textbf{Notations.}\:
An attributed graph can be denoted as $G=(V,E, \boldsymbol{X}, \boldsymbol{Y})$, where $V$ and $E$ are the sets of nodes and edges respectively, $\boldsymbol{X}\in \mathbb{R}^{|V|\times f}$ is the node feature matrix, and $\boldsymbol{Y}\in \mathbb{R}^{|V|\times C}$ is the node label matrix. 
Here we use $|V|$, $f$, $C$ to denote the number of nodes, the dimension of the node features, and the number of node classes, respectively.
Each row of $\boldsymbol{X}$ is the feature vector of node $v_i$, and each row of $\boldsymbol{Y}$ is the one-hot label of node $v_i$.
The structure elements $(V, E)$ can also be denoted as an adjacency matrix $\boldsymbol{A}\in\mathbb{R}^{|V|\times|V|}$ that encodes pairwise connections between the nodes, whose entry $\boldsymbol{A}_{ij} = 1$ if there exists an edge between $v_i$ and $v_j$, and $\boldsymbol{A}_{ij} = 0$ otherwise.
Based on the adjacency matrix, we can define the degree distribution of $G$ as a diagonal degree matrix $\boldsymbol{D} \in \mathbb{R}^{|V|\times |V|}$ with entries $\boldsymbol{D}_{ii} = \sum_{j=1}^{|V|} \boldsymbol{A}_{ij} = d_i$ representing the degree value of $v_i$.

\noindent\textbf{Graph Convolutional Network.}\:
GCN~\cite{kipf2016semi} can be formulated as: 
\begin{equation}
    \boldsymbol{B} = \operatorname{softmax}\left(\widetilde{\boldsymbol{A}} \cdot \operatorname{ReLU}\left(\widetilde{\boldsymbol{A}} \boldsymbol{X} \boldsymbol{W}_0\right) \cdot\boldsymbol{W}_1\right)
\end{equation}
where the output $\boldsymbol{B}$ is the soft assignment of label predictions, and GCN-style regularization is:
\begin{equation}\label{equ:regular}
    \widetilde{\boldsymbol{A}}=(\boldsymbol{D}+\boldsymbol{I})^{-1/2}(\boldsymbol{A}+\boldsymbol{I})(\boldsymbol{D}+\boldsymbol{I})^{-1/2}
\end{equation}
 
% GCN can be trained by minimizing cross-entropy loss under semi-supervised learning settings.

\begin{figure}[htp]
  \begin{minipage}[t]{0.49\linewidth}
      \centering
      \includegraphics[width=\linewidth]{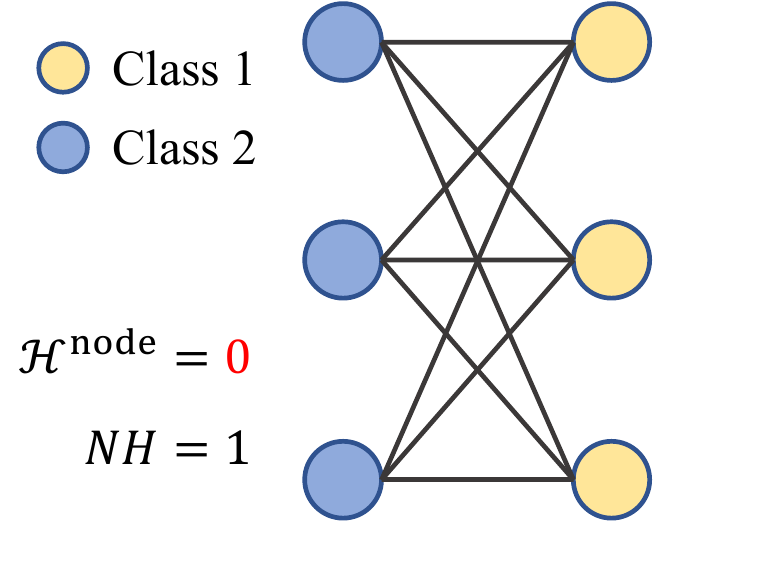} 
      \captionsetup{margin=5pt}
      % \caption{Case shows that \textit{NH} can measure the prediction ability of GCN but $\mathcal{H}^{node}$ metric fails.}
      \caption{\textit{NH} can measure the prediction ability of GCN but $\mathcal{H}^{node}$ fails.}
      \label{fig:demo}
  \end{minipage}
      \begin{minipage}[t]{0.49\linewidth} %所有minipage宽度之和要小于1，否则会自动变成竖排
          \centering
          \includegraphics[width=\linewidth]{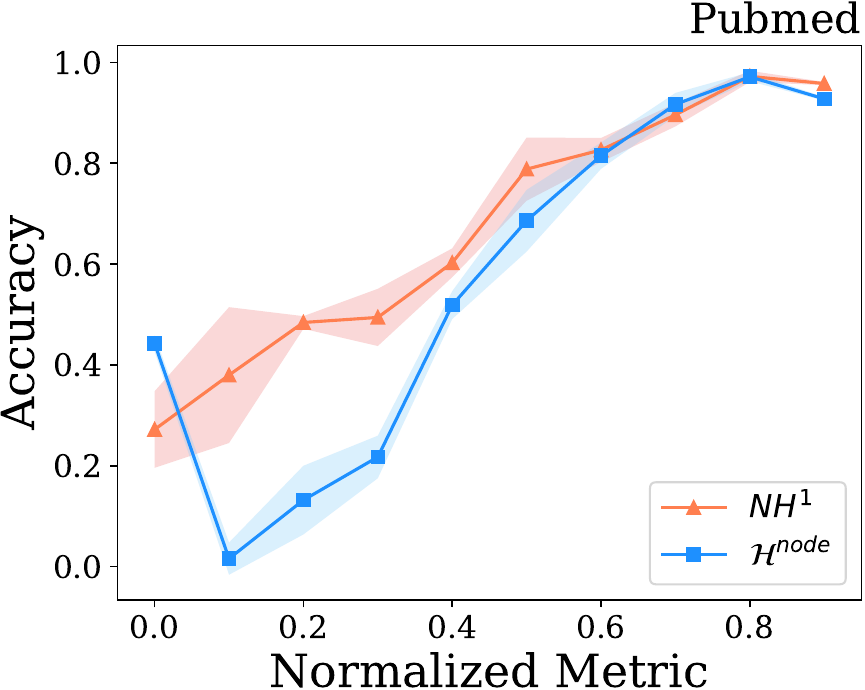}
          \captionsetup{margin=5pt}
          \caption{Correlation between accuracy of GCN and two metrics.}
          \label{fig:metric_acc}
      \end{minipage}
  \end{figure}

\noindent\textbf{Node Homophily.}\:
\label{sec:Metrics of Homophily}
This metric is proposed in~\cite{pei2019geom} and is widely used to measure the homophily, which can be defined as follows:
\begin{equation}
    \begin{aligned}
        &\text{node-level:} \quad  \mathcal{H}_i^\text{node}=\frac{1}{d_i} \cdot {|\{v_j \mid (v_i,v_j)\in E,y_i=y_j\}|}  \\
        &\text{graph-level:} \quad  \mathcal{H}_G=\frac{1}{|V|}\sum_{v_i \in V}{\mathcal{H}_i^\text{node}}
    \end{aligned}
\end{equation}
Node-level homophily denotes the proportion of direct neighbors that share the same label as the target node, whereas graph-level metric evaluates global homophily of a graph by averaging across all nodes.
Most existing studies utilize graph-level homophily to distinguish between homophilous and heterophilous graphs, yet they overlook the significance at node level.
Zhu et al.~\cite{zhu2020h2gcn} first observed the \textit{tick}-like phenomenon between node-level homophily and accuracies (See Fig.~\ref{fig:metric_acc}). 
UDGNN~\cite{liu2022ud} first suggested that nodes with low or high homophily should be distinguished using the metric within one graph. Nevertheless, this classical metric falls short in many recent observations and is restricted to first-order neighborhoods, making it difficult to include more label information.

%%%%%%%%%%%%%%%%%%%%%%%%%%%%%%%%%%%%%%%%%%%%%%%%%%%%%%%%%%%%%%%%%%%%%%%%%%%% section 3 %%%%%%%%%%%%%%%%%%%%%%%%%%%%%%%%%%%%%%%%%%%%%

\section{Methodology}
\subsection{Neighborhood Homophily Metric}
Generally, heterophily is not conducive to classical GNNs because the features of different classes of nodes will be inappropriately mixed during message aggregation, leading to the learned node features being indistinguishable~\cite{zhu2020h2gcn}.
However, GCN still performs well on bipartite graphs that are completely heterophilous under the definition of the node homophily metric, bringing contradictory judgments~\cite{ma2021necessity}.
As illustrated in Fig.~\ref{fig:demo}, node homophily ($\mathcal{H}^\text{node} = 0$) cannot explain the fact that GCN still performs well on bipartite graphs.
To overcome the shortcomings of classical homophily metrics, we propose a new metric termed Neighborhood Homophily (\textit{NH}), which can measure the label complexity or purity in the neighborhoods of target nodes.

% \subsubsection{Definition}
\noindent\textbf{Definition.}\:
For a target node $v_i$, its $k$-hop neighborhood homophily $\textit{NH}_i^k$ can be defined as follows:
% \begin{equation}
%     NH_i^k=\frac{1}{d_i^k} \cdot {\mathop{max}\limits_{j \in h^k(i)}\{count(y_j)\}}\label{equ:CC}     
% \end{equation}
\begin{equation}\label{eq: NH}
  \textit{NH}_i^k= \  \frac{|\mathcal{N}(i,k,c_\textit{max})|}{|\mathcal{N}(i,k)|} \quad \text{with} \quad c_\textit{max} =\underset{c \in [1,C]}{\arg \max }\ \  |\mathcal{N}(i,k,c)| 
\end{equation}
where $ \mathcal{N}(i,k) =\{v \mid  (0 \: \text{or} \: 1)\leq  \operatorname{ShortestPath}(v_i,v) \leq k \} $ means the neighbor set in the $k$-hop neighborhood of $v_i$, $\mathcal{N}(i,k,c) = \{v_j \mid v_j \in \mathcal{N}(i,k), y_j = c \}$ means the set of neighbors whose node label is $c$ in the $k$-hop neighborhood of $v_i$. The hop $k$ and whether $\mathcal{N}(i,k)$ includes the target node $v_i$ (0 or 1) are treated as hyper-parameters.
For an isolated node, we set its \textit{NH} value to 1. 
% Obviously, $\textit{NH} \in [1/C, 1] $ considering the most complex and pure neighborhood.
the value domain of \textit{NH} can be determined as $\textit{NH} \in [1/C, 1]$, where the minimum and maximum values correspond to the most complex and purest neighborhoods, respectively.

In Fig.~\ref{fig:demo}, $\textit{NH}=1$ indicates that the \textit{NH} metric considers the neighborhoods of the nodes in the bipartite graph to have the lowest complexity (or the highest purity) and will not confuse GCN.

\noindent\textbf{Comparison.}\:
Since classical GNNs such as GCN are designed based on homophily assumption, they naturally perform poorly in low homophily scenarios. Therefore, a well-designed homophily metric should behave consistently with the performance of GCN.
To facilitate comparison between \textit{NH} and the node homophily, we normalize these two metrics to unify their ranges to $[0,1]$.
Utilizing these two metrics, we divide all nodes into ten groups according to their metric levels, ranging from 0 to 1 in intervals of 0.1. The ordinate represents the intra-group accuracies on the standard GCN (2 layers with 64 hidden units, 0.5 dropout rate, and ReLU.).
As shown in Fig.~\ref{fig:metric_acc}, the accuracy curve represented by \textit{NH} rises almost monotonically, reflecting that GCN performs better for prediction in scenarios with higher \textit{NH}. 
However, GCN has abnormally high prediction accuracy for the nodes with the lowest node homophily.
Such phenomena well illustrates that \textit{NH} metric can better measure the difficulty of classifying nodes with GCN, i.e., \textit{NH} is a better metric than node homophily.
Meanwhile, we also compare the prediction results of our proposed model and GCN on different node groups, as shown in Fig.~\ref{fig:model_acc}, from which we can see that with the guidance of \textit{NH} metric, our model can better predict low homophily nodes while maintaining the prediction effect on high homophily nodes when compared with GCN.

%%%%%%%%%%%%%%%%%%%%%%%%%%%%%%%%%%%%%%%%%%%%%%%%%%%%%%%%%%%%%%%%%%%%%%

\subsection{Neighborhood Homophily Mask}\label{sec: NH-mask}
The key design of \textit{NH} metric is that it counts the label distribution, which is target-driven in the case of semi-supervised learning. Moreover, label distribution is the direct reason why graphs are categorized as homophilous or heterophilous. 
During the training phase, we follow a valid and common practice~\cite{zhu2021cpgnn}, that is, using the predicted labels partly complemented by ground-truth labels of the training set.
% During the training phase, we follow a valid and common practice~\cite{zhu2021cpgnn}, that is, replacing the ground-truth labels of the training set with predicted labels. 
We first initialize the \textit{NH} values of all nodes to 1, and then update the \textit{NH} values at each subsequent epoch when the validation accuracy achieves a new high.
Each time the \textit{NH} value are updated, we utilize a threshold $T$ to separate the nodes into low- and high-\textit{NH} ones, thereby enabling us to generate the Neighborhood Homophily Masks (\textit{NH} mask):
\begin{equation}\label{eq: mask}
    \boldsymbol{m}^\text{low}_{i} = \mathbb{I}(\textit{NH}_i^k < T)\ , \quad \boldsymbol{m}^\text{high}_{i} = \mathbb{I}(\textit{NH}_i^k \geq T) 
\end{equation}
where $\boldsymbol{m}^\text{low}$ (or $\boldsymbol{m}^\text{high}$) $\in \mathbb{R}^{1 \times |V|}$ is mask vector. The absolute separation of nodes by setting threshold ensures the hard topology optimization, also used in~\cite{zou2023similarity,klicpera2019diffusion}.

\begin{figure}[htbp]
  \centering %图片局部居中
  \includegraphics[width=\linewidth]{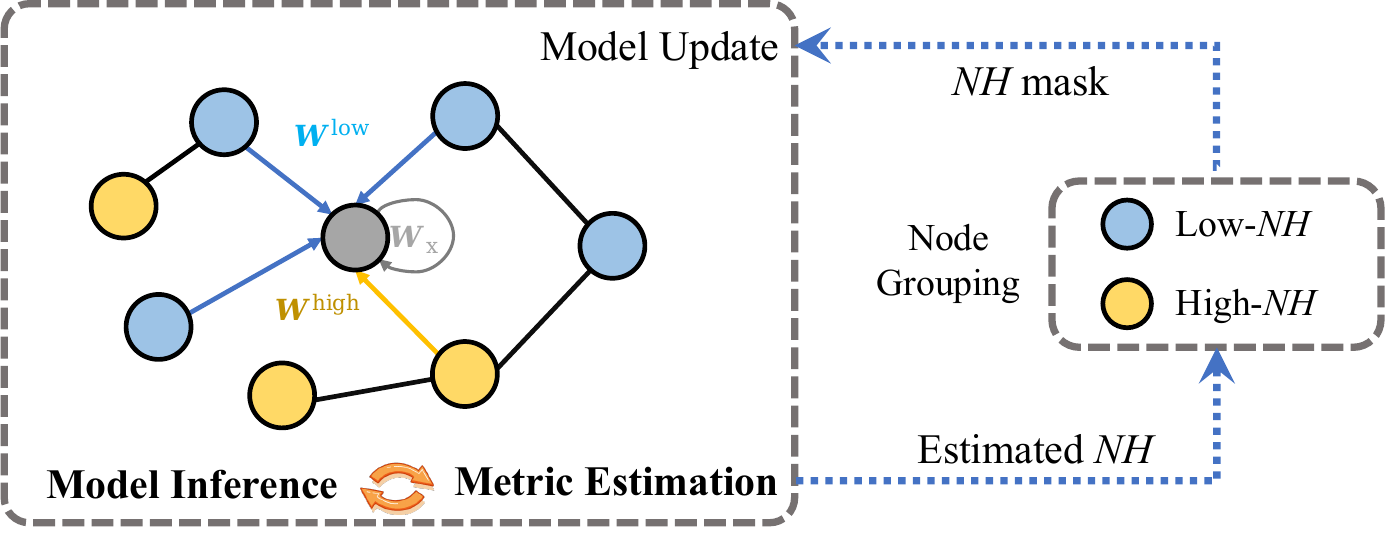}
  \caption{Illustration of NHGCN framework. Two alternating processes are: 1) each time the validation accuracy achieves a new high, estimate the \textit{NH} values of nodes and update the \textit{NH} masks, which can divide all nodes into two groups (low- and high-\textit{NH}); 2) the two groups of nodes are asymmetrically aggregated in two different channels, and the raw node features are mapped in the third channel, and the outputs of the three channels are combined for subsequent model training.}
  \label{fig:main}
\end{figure}

\subsection{NH-based Message Propagation}
% \subsection{Neighborhood Homophily-based Message Propagation}
As mentioned above, low- and high-\textit{NH} nodes have different behavior patterns.
Therefore, during message propagation, the low- and high-\textit{NH} neighbors around target nodes should also transmit their messages from different channels.
% From the perspective of the central node during message aggregation, these two kinds of neighboring nodes should be aggregated from different channels. 
Here we design \textbf{N}eighborhood \textbf{H}omophily-based \textbf{G}raph \textbf{C}onvolutional \textbf{N}etwork (\textbf{NHGCN}), which utilizes the \textit{NH} masks to guide the message propagation.
The backbone of NHGCN is a mask-based deformation of GCN architecture:
\begin{equation}\label{eq: layer}
    \boldsymbol{H}^s = \widetilde{\boldsymbol{A}}_\text{mask} \cdot \operatorname{ReLU}\left(\widetilde{\boldsymbol{A}}_\text{mask} \cdot \boldsymbol{X} \boldsymbol{W}_0^s\right) \cdot \boldsymbol{W}_1^s\ , \quad \boldsymbol{A}_\text{mask}=\boldsymbol{A}\odot\boldsymbol{m}^s
\end{equation}
where $\boldsymbol{W}_0^{s}\in \mathbb{R}^{f\times f'}$ and $\boldsymbol{W}_1^{s} \in \mathbb{R}^{f'\times f'}$ are the weight matrices, $f'$ is the hidden dimension, and $s\in\{\text{low}, \text{high} \}$ indicates different learning channels. 
Note that $\boldsymbol{A}\odot\boldsymbol{m}^s$ implies performing row-wise Hadamard product on $\boldsymbol{A}$, 
which separates neighbors into two channels and adopt channel-specific weights. Then the masked adjacent matrix is regularized in a GCN-style as Equ.~\ref{equ:regular}.
Since the existence of isolated and neighbor-independent nodes, we introduce a third channel to preserve the raw features:
\begin{equation}
    \boldsymbol{H}_x =\operatorname{ReLU}\left(\boldsymbol{X}\boldsymbol{W}_\text{x}\right)
\end{equation}
where $\boldsymbol{W}_\text{x} \in \mathbb{R}^{f\times f'}$ is the weight matrix.
Finally, we combine the representations from the three channels and derive the soft assignment prediction $ \boldsymbol{B} \in \mathbb{R}^{n\times C}$ through a MLP as follows:
\begin{equation}
        \boldsymbol{B}  = \operatorname{softmax}\left(\left(\alpha_\text{low}\boldsymbol{H}^\text{low}+\alpha_\text{high}\boldsymbol{H}^\text{high}+\alpha_\text{x} \boldsymbol{H}_\text{x}\right)\cdot \boldsymbol{W}_\text{o}\right)
\end{equation}
where $\mathbf{W}_\text{o} \in \mathbb{R}^{f'\times C}$ is the weight matrix in the MLP, and $\alpha_\text{low}+\alpha_\text{high}+\alpha_\text{x}=1$, where $\alpha$ are learnable scale weights initialed by $\frac{1}{3}$. Finally, we employ the cross-entropy as the classification loss.

\begin{table*}
  \centering
  \caption{Results on real world benchmark datasets: Mean Test Accuracy (\%) $\pm$ Standard Deviation. The best result on each dataset is marked in bold while the second best is underlined. Time consumption implies the average time per run on Cora.}
  \label{tab:main}
  \resizebox{\textwidth}{!}{
  \renewcommand{\arraystretch}{1.1}
  \begin{tabular}{cccccccccccr} 
  \hline
                     & Cora                  & Citeseer              & Pubmed                & Photo                 & Computers             & Actor                 & Chameleon*            & Squirrel*             & Cornell               & Texas                 & \multicolumn{1}{c}{Time/s}  \\ 
  \hline
  \textit{MLP}       & 73.74 $\pm$ 1.73          & 76.85 $\pm$ 1.11          & 86.86 $\pm$ 0.51          & 91.28 $\pm$ 0.63          & 84.35 $\pm$ 1.02          & 36.82 $\pm$ 1.19          & 31.99 $\pm$ 4.11          & 44.61 $\pm$ 4.23          & 80.98 $\pm$ 7.01          & 82.34 $\pm$ 5.12          & \textit{0.56}               \\
  \textit{GCN}       & 87.50 $\pm$ 1.56          & 81.11 $\pm$ 1.06          & 88.08 $\pm$ 0.50          & 93.94 $\pm$ 0.47          & 89.72 $\pm$ 0.43          & 33.15 $\pm$ 1.75          & 42.63 $\pm$ 3.64          & 47.83 $\pm$ 2.90          & 79.18 $\pm$ 3.94~         & 88.24 $\pm$ 1.50          & \textit{0.37}               \\ 
  \hline
  \textit{GPRGNN}    & \textbf{89.46 $\pm$ 1.40} & 81.84 $\pm$ 1.08          & \underline{90.46 $\pm$ 0.75}  & 94.62 $\pm$ 0.45          & 89.35 $\pm$ 1.33          & 39.22 $\pm$ 1.17          & \underline{45.15 $\pm$ 4.95}  & 45.85 $\pm$ 2.33          & 90.66 $\pm$ 4.58~         & 88.72 $\pm$ 6.50          & \textit{1.35}               \\
  \textit{ACMGCN}    & 87.80 $\pm$ 1.23          & \textbf{82.05 $\pm$ 1.26} & 90.40 $\pm$ 0.29          & \underline{95.04 $\pm$ 0.44}  & \underline{90.31 $\pm$ 0.39}  & 39.09 $\pm$ 1.42          & 43.74 $\pm$ 4.03          & 39.64 $\pm$ 7.84          & \textbf{91.48 $\pm$ 2.29} & 88.09 $\pm$ 3.20          & \textit{0.96}               \\
  \textit{FAGCN}     & 88.90 $\pm$ 0.88~         & 80.78 $\pm$ 0.95          & 89.16 $\pm$ 0.67          & 94.89 $\pm$ 0.48          & 87.61 $\pm$ 1.17          & \underline{40.90 $\pm$ 1.12}  & 45.20 $\pm$ 4.45          & \underline{53.28 $\pm$ 1.28}  & 88.52 $\pm$ 4.95~         & \underline{89.57 $\pm$ 3.94}  & \textit{1.60}               \\ 
  \hline
  \textit{GGCN}      & 87.34 $\pm$ 1.15          & 78.95 $\pm$ 1.77          & 89.03~ $\pm$ 0.43         & 91.26 $\pm$ 0.80          & 75.98 $\pm$ 7.68          & 38.86 $\pm$ 0.80          & 40.70 $\pm$ 4.00          & 49.31 $\pm$ 4.62          & 90.66 $\pm$ 3.28          & 83.40 $\pm$ 6.91          & \textit{6.18}               \\
  \textit{GBK}       & 88.69 $\pm$ 0.42          & 79.18 $\pm$ 0.96          & 85.49 $\pm$ 0.61          & 81.61 $\pm$ 8.05          & 81.20 $\pm$ 5.62          & 37.27 $\pm$ 0.98          & 45.32 $\pm$ 4.61          & 45.41 $\pm$ 3.19          & 81.08 $\pm$ 4.88          & 74.27 $\pm$ 2.18          & \textit{105.08}             \\
  \textit{WRGAT}     & 88.20 $\pm$ 2.26          & 76.81 $\pm$ 1.89          & 89.13 $\pm$ 0.53          & 92.23 $\pm$ 0.64          & 86.13 $\pm$ 0.76          & 37.91 $\pm$ 1.63          & 42.46 $\pm$ 3.71          & 45.47 $\pm$ 3.56          & 83.62 $\pm$ 5.50          & 81.62 $\pm$ 3.90          & \textit{25.39}              \\ 
  \hline
  \textit{NHGCN}     & \underline{89.05 $\pm$ 1.24}  & \underline{81.87 $\pm$ 1.20}  & \textbf{91.56 $\pm$ 0.50} & \textbf{95.35 $\pm$ 0.34} & \textbf{90.72 $\pm$ 0.34} & \textbf{43.94 $\pm$ 1.14} & \textbf{47.60 $\pm$ 4.15} & \textbf{54.38 $\pm$ 1.17} & \underline{91.28 $\pm$ 3.94}  & \textbf{93.11 $\pm$ 2.42} & \textit{3.68}               \\
  \textit{w/o group} & 88.62 $\pm$ 1.70          & 81.31 $\pm$ 1.10          & ~90.09 $\pm$ 0.92~        & 94.67 $\pm$ 0.33          & 89.21 $\pm$ 0.77          & 39.67 $\pm$ 0.75          & 41.17 $\pm$ 7.21          & 53.28 $\pm$ 2.43          & 89.36 $\pm$ 4.60          & 87.23 $\pm$ 4.26          & \textit{3.33}               \\
  \hline
  \end{tabular}}
\end{table*}

\section{Evaluations}
% \subsection{Datasets}
\subsection{Experiment Setup}
\noindent\textbf{Datasets.}\:
We use 10 real-world datasets as in~\cite{chien2020adaptive}, among which the first five (\textit{Cora}, \textit{Citeseer}, \textit{Pubmed}, \textit{Computers} and \textit{Photo}) are homophilous datasets and the last five (\textit{Chameleon*}, \textit{Squirrel*}, \textit{Actor}, \textit{Cornell} and \textit{Texas}) are heterophilous ones. Specifically, \textit{Chameleon*} and \textit{Squirrel*} are de-duplicated version to get rid of the risk of data leakage provided by \cite{critical}.
We adopt dense split as~\cite{chien2020adaptive}, i.e., 60\% / 20\% / 20\% for training / validation / testing set. 

\noindent\textbf{Baselines.}\:
We compare NHGCN with three categories of eight baselines: 
1) basic methods: two-layer MLP, GCN~\cite{kipf2016semi}; 
2) spectral methods: GPRGNN~\cite{chien2020adaptive}, ACMGCN~\cite{luan2022acm}, FAGCN~\cite{bo2021fagcn};
3) spatial methods: GGCN~\cite{yan2022two}, GBK~\cite{du2022gbk}, WRGAT~\cite{suresh2021wrgat}.

\noindent\textbf{Settings.}\:
We report the average results and running time of 10 runs fixed by 10 random seeds in Table~\ref{tab:main}. For all baselines, we reproduce them with their open-sourced code.
For all methods, we set optimizer to Adam~\cite{kingma2014adam}, hidden dimension to 512, the maximum epochs to 500 with 100 epochs patience for early stopping. The search space for learning rate, weight decay and dropout rate will be shared among all methods. We only have specific hyper-parameters $ k $ chosen from \{1, 2\}, \textit{if\_include\_self} when calculate \textit{NH}, and the threshold $T$ chosen from $[1/2, 1/C]$ with 0.5 internals for the reciprocal (maximal is set to 4). 
The source code of NHGCN is available at \url{https://github.com/rockcor/NHGCN}.

\subsection{Analysis}
\noindent\textbf{Node Classification.}\:
Table~\ref{tab:main} reports the node classification results on ten benchmarks, from which we can observe and conclude:
1) Compared with eight latest competitors, our NHGCN shows top average ranking, indicating that our methods are effective and have better universality for datasets with different level of homophily. More specifically, our method achieves SOTA on 7 out of 10 benchmarks with relatively low time consumption and variance, guaranteeing the efficiency and stability. In particular, we achieved 7.4\% improvement on the \textit{Actor}, significantly higher than all existing GNNs to our best knowledge.
2) Baselines show preferences for different datasets. Generally, spectral methods performs better on datasets with high homophily, while spatial methods performs better on datasets with high heterophily. In addition, most spatial methods run with an unacceptable time consumption.

\noindent\textbf{Ablations.}\:
When our grouping strategy is removed, NHGCN degenerates to a combination of GCN output and raw features, represented by \textit{w/o group} for short. Other components such as layers and combiner is the same as NHGCN. The results are shown in the last row of Tab.~\ref{tab:main}. When comparing NHGCN with \textit{w/o group}, it is well illustrated that treat neighbors differently is an effective scheme.

\begin{figure}[htp]
  \begin{minipage}[t]{0.49\linewidth} %所有minipage宽度之和要小于1，否则会自动变成竖排
    \centering %图片局部居中
    \includegraphics[width=\linewidth]{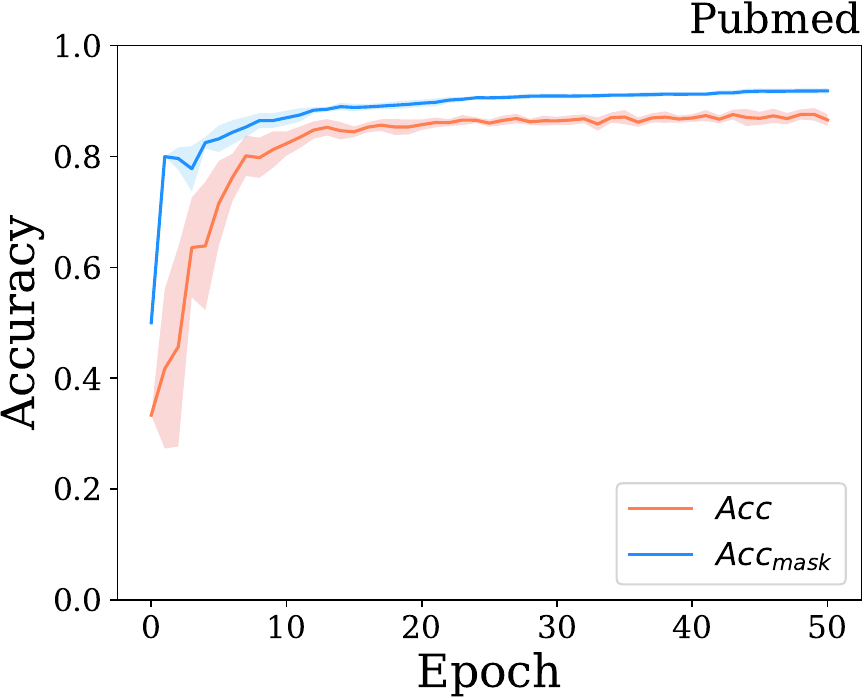}
    \captionsetup{margin=4pt}
    % \caption{Accuracy curves for node classification and \textit{NH} masks.}
    \caption{Accuracy of node classification and masks.}
    \label{fig:metric analysis}
  \end{minipage}
  \begin{minipage}[t]{0.49\linewidth} %所有minipage宽度之和要小于1，否则会自动变成竖排
    \centering%图片局部居中
    \includegraphics[width=\linewidth]{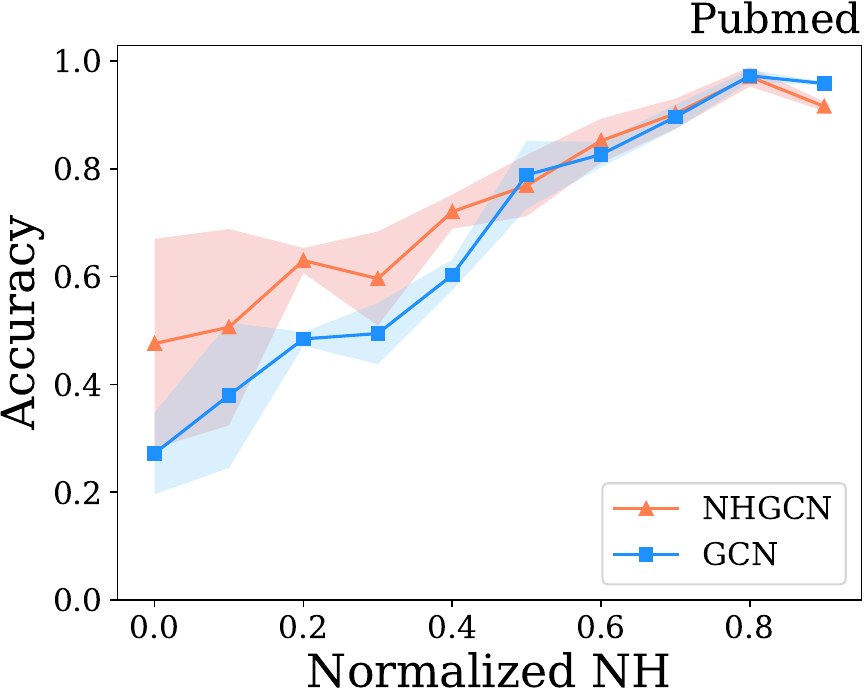}%此时的图片宽度比例是相对于这个minipage的，不是全局
    \captionsetup{margin=4pt}
    % \caption{Accuracy of GCN and NHGCN in different \textit{NH} levels.}
    \caption{Accuracy of two models in varying \textit{NH}.}
    \label{fig:model_acc}
  \end{minipage}
\end{figure}

\noindent\textbf{Metric Effectiveness.}\:
Our grouping strategy is totally based on the \textit{NH} metric. However, the \textit{NH} is calculated by the predicted labels. To find out if \textit{NH} really plays a role in our model, and if the co-optimization achieved, we plot the accuracies of the estimated \textit{NH} along with the node classification, as shown in Fig.~\ref{fig:metric analysis}. During training, the two lines rise in synchrony, proving that the two accuracies are interdependence and the two processes are co-optimized.

\noindent\textbf{Debiasing Potential.}\:
Fig.~\ref{fig:model_acc} shows that NHGCN surpasses GCN at almost all \textit{NH} levels, thus it can achieve better overall accuracy on \textit{Pubmed}. This suggests that our grouping strategy benefits low-\textit{NH} nodes more than high-\textit{NH} ones.
From perspective of debiasing learning~\cite{liu2022ud}, it significantly reduces the accuracy gap between low- and high-\textit{NH} nodes, ensuring that classification performance is relatively maintained even if the proportion of nodes in these two groups changes.
In practice, this also means that our model is fair to both ``outward'' and ``inward'' nodes. 
Such phenomena are also reflected in other datasets.

\section{Conclusions}
In this paper, we address two common problems of GNNs tailed-made for graph universality problem: 1) The model training process is not well associated with their proposed metrics; 2) High time consumption for both model training and metric calculating.
We design neighborhood homophily, as an easy-to-compute metric to measure the homophily, which overcomes the shortness of classical metrics.
By a slight transformation of GCN, our metric can guide message propagation in a grouping strategy, simply but effectively.


\begin{acks}
  This work was partially supported by the Key R\&D Programs of Zhejiang, China under Grant No. 2022C01018, by the Zhejiang Provincial Natural Science Foundation of China under Grants No. LR19F030001, and by the National Natural Science Foundation of China under Grant No. 61973273.
\end{acks}

\bibliographystyle{ACM-Reference-Format}
% \balance
\bibliography{sample-base}

% \appendix

% \newpage

% \begin{multicols}{2}
% \onecolumn
% \input{appendix.tex}
% \end{multicols}

\end{document}